%% file: main.tex
\documentclass{article} 
\usepackage{iclr2027_conference,times}

\input{math_commands.tex}

\usepackage{url}

\definecolor{iccvblue}{rgb}{0.21,0.49,0.74}
\usepackage[colorlinks,linkcolor=red,citecolor=iccvblue]{hyperref}
\iclrfinalcopy
\usepackage{multirow}
\usepackage{colortbl}
\usepackage{graphicx}
\usepackage{booktabs}
\usepackage{xcolor}
\usepackage{verbatim}
\usepackage{arydshln}
\usepackage{array}
\usepackage{makecell}
\usepackage{caption}
\usepackage{wrapfig}
\usepackage{subcaption}
\usepackage{amssymb}
\usepackage{pifont}
\usepackage{enumitem}

\newcommand{\eg}{\textit{e.g.}}

\definecolor{LoopTrackHeader}{RGB}{240,240,245}

\definecolor{LoopTrackHighlight}{RGB}{224,244,247}

\newcommand{\tablegroup}[1]{%
  \multicolumn{15}{l}{%
    \cellcolor{LoopTrackHeader}\textbf{#1}%
  }\\%
}

\title{LoopTrack: A Simple Baseline for\\ Parameter-Efficient Transformer Tracking}

\author{%
\parbox[t]{\dimexpr\textwidth-2\tabcolsep\relax}{%
\centering
\bfseries
Liang Peng$^{1}$\thanks{%
Equal contribution.\quad
$^{\dagger}$Primary corresponding author.\quad
$^{\ddagger}$Secondary corresponding author.%
}
\quad
Chenxiao Li$^{2,*}$
\quad
Libo Zhang$^{3}$
\quad
Xingping Dong$^{1,\dagger}$
\quad
Heng Fan$^{2,\ddagger}$
\\[5pt]
\normalfont
$^{1}$School of Computer Science, National Engineering Research Center for Multimedia Software,
\\
Institute of Artificial Intelligence, Hubei Key Laboratory of Multimedia and Network Communication Engineering, Wuhan University
\\[3pt]
$^{2}$University of North Texas
\qquad
$^{3}$Institute of Software, Chinese Academy of Sciences
\\[4pt]
\texttt{pengliang@whu.edu.cn}
}}

\begin{document}




\maketitle
\fancyhead{}
\renewcommand{\headrulewidth}{0pt}

\begin{abstract}
Current Transformer-based tracking methods typically stack multiple Transformer blocks with separate parameters to model interactions between the target template and the search region for target localization. Despite excellent performance, these trackers often incur substantial parameter overhead from stacked blocks, making their deployment on resource-limited devices difficult. To address this, we propose a novel parameter-efficient Transformer tracking framework, dubbed \textbf{\emph{LoopTrack}}, which repeatedly applies a small set of Transformer blocks with shared parameters to interact features in a looped architecture for tracking, significantly reducing the number of parameters. To further exploit target cues for improving LoopTrack, we present two lightweight designs, including \emph{target-aware looping} (TAL) and \emph{gated target memory} (GTM). The former applies intermediate target information generated by one loop to guide feature interaction in the subsequent loop, enabling progressive feature refinement, while the latter maintains a compact memory across frames, which is incorporated into the loop process to provide long-term information to the tracker, mitigating temporal drift in tracking. Compared to existing Transformer trackers, LoopTrack enables multiple rounds of feature interaction with fewer model parameters, making it resource-friendly for deployment. In extensive experiments on multiple datasets, LoopTrack shows promising results with a favorable accuracy-parameter trade-off. In particular, our LoopTrack$_{\text{One}}$, with a single shared Transformer block, achieves 66.2\% SUC score on LaSOT with only 3.4M parameters, while LoopTrack$_{\text{Three}}$, using three shared blocks, achieves 69.3\% SUC score with 6.4M model parameters, surpassing existing parameter-efficient tracking methods with comparable or larger model size. With LoopTrack, we aim to establish a simple yet strong baseline for parameter-efficient Transformer tracking. Our code and models will be released.
\end{abstract}

\section{Introduction}
\label{sec:introduction}

Visual tracking, aiming to localize the target of interest in a video given its initial box, has been extensively studied in the past decades owing to its crucial roles in many applications, such as robotics, video surveillance, and autonomous vehicles. By embracing the Transformer architecture~\citep{VaswaniSPUJGKP17,DosovitskiyB0WZ21}, visual tracking has witnessed considerable progress in recent years~\citep{gao2022aiatrack,chen2021transformer,yan2021learning,gao2023generalized,lin2022swintrack}. In particular, the one-stream Transformer tracking paradigm~\citep{ye2022joint}, with its simple architecture and excellent performance, has attracted extensive attention from the tracking community and become the dominant Transformer tracking framework~\citep{chen2023seqtrack,wei2023autoregressive,lin2024tracking,lin2026loratv2}. These trackers typically stack multiple Transformer
blocks to model interactions between the target template and search region for target localization. Since each Transformer block in these methods is independently parameterized, stacking multiple blocks, often 18, 24, or even 40, depending on the backbone, can lead to substantial parameter overhead and rapidly increases model size, making deployment on resource-constrained devices increasingly challenging.

To alleviate the parameter burden of Transformer trackers, existing parameter-efficient\footnote{It is worth noting that, here parameter-efficient refers to reducing the number of model parameters required for tracking, rather than parameter-efficient fine-tuning.} approaches usually construct a compact student model through network compression~\citep{hong2025general}, depth pruning~\citep{cui2023mixformerv2}, or architectural reduction~\citep{kang2026uetrack}, and then apply knowledge distillation~\citep{hinton2015distilling} from a stronger teacher to improve performance. Although effective, this distillation-based paradigm has several limitations. First, distillation relies on an additional strong teacher tracker, which typically needs to be pretrained or otherwise available, thereby introducing more complexity with additional computational and supervision overhead. Second, obtaining a highly compact tracker, \eg, with fewer than 10M parameters, usually requires substantially reducing the number or capacity of Transformer blocks, which also decreases the feature interaction in Transformer. While distillation can compensate for part of this loss, preserving strong tracking performance becomes increasingly difficult under such a small parameter budget. 

To mitigate these limitations in existing approaches, we explore parameter sharing as an alternative solution to building compact Transformer tracking models. Parameter sharing in Transformer~\citep{dehghani2018universal,shen2022sret} aims at reusing the same set of parameters across multiple processing stages for representation learning, rather than allocating independent parameters to each stage. Drawing inspiration from this, we propose to decouple feature interactions between template and search region from the number of independently parameterized Transformer blocks. Instead of assigning a separate set of parameters to each Transformer block, we repeatedly reuse a small set of shared Transformer blocks, substantially reducing the number of model parameters for tracking. 

Based on this idea, we propose \textbf{\emph{LoopTrack}}, a novel parameter-efficient Transformer tracking framework which applies parameter-shared Transformer blocks within a recurrent loop to interact features for target localization. In LoopTrack, the output feature from one loop are fed into the same shared Transformer blocks in the subsequent loop, enabling multiple rounds of feature interaction between the target template and the search region, as in conventional Transformer trackers, while requiring substantially fewer model parameters. While reducing model parameters, the basic looped design does not explicitly exploit target-specific cues during recurrent interaction. To further enhance LoopTrack with target cues, we present two compact designs, comprising \emph{target-aware looping} (TAL) and \emph{gated target memory} (GTM). TAL leverages intermediate target localization response generated in one loop through a lightweight prediction head to guide feature interaction in the subsequent loop, enabling progressive refinement of the learned representations for tracking. GTM maintains a compact target memory across frames, which is updated with reliable prediction result in each frame and incorporated into the loop process to
provide long-term information to the tracker, mitigate tracking drift. Together, TAL and GTM exploit complementary target cues within and across frames to enhance LoopTrack, while introducing negligible additional parameters and computational overhead. 


In comparison to conventional Transformer trackers with independently parameterized Transformer blocks, LoopTrack enables feature interaction with fewer model parameters for localization, making it more resource-friendly for deployment. Besides, unlike existing parameter-efficient Transformer trackers, LoopTrack does not rely on an additional teacher model for distillation while maintaining a compact architecture with strong tracking accuracy, leading to a simpler training pipeline. In order to validate the effectiveness of LoopTrack, we conduct extensive experiments on four tracking benchmarks, including LaSOT~\citep{fan2019lasot},
LaSOT$_{\mathrm{ext}}$~\citep{fan2021lasot}, TrackingNet~\citep{muller2018trackingnet}, and
GOT-10k~\citep{huang2021got10k}. Our LoopTrack$_{\text{Three}}$, with three shared Transformer blocks, achieves 69.3\% SUC on LaSOT using only 6.4M parameters, delivering accuracy comparable to that of the parameter-efficient tracker UETrack-B~\citep{kang2026uetrack} at approximately half its parameter count. Even our lighter LoopTrack$_{\text{One}}$, using a single shared Transformer block, outperforms UETrack-T by 2.8 percentage points in SUC, reaching 66.2\% with only 3.4M parameters 43\% fewer than UETrack-T. Notably, LoopTrack can operate at different inference depths simply by varying the number of loops, allowing the same trained
model to balance tracking accuracy and speed without introducing additional model parameters.

In summary, our \textbf{contributions} are as follows:
\ding{171} We propose LoopTrack, a parameter-efficient Transformer tracking framework. It substantially reduces model parameters through looped parameter sharing while maintaining strong tracking performance;
\ding{170} We propose TAL and GTM, two lightweight designs. They enable the shared Transformer blocks to use intermediate localization cues and cross-frame target information for more effective feature refinement;
\ding{169} Extensive experiments show that LoopTrack achieves consistent performance improvements over other parameter-efficient tracking methods and offers a favorable accuracy-parameter trade-off.

\section{Related Work}
\label{sec:related_work}

\paragraph{Transformer-based Tracking.}
Early Transformer-based tracking methods, such as TransT~\citep{chen2021transformer} and STARK~\citep{yan2021learning}, employ attention modules to model relations between template and search features extracted by a backbone.
Subsequent methods, represented by MixFormer~\citep{cui2022end} and OSTrack~\citep{ye2022joint}, unify feature extraction and template-search interaction within a one-stream architecture, a formulation adopted by SUTrack~\citep{chen2025sutrack}.
This integration simplifies the tracking pipeline and enables feature refinement through successive attention blocks.
However, these architectures typically allocate separate parameters to blocks at different depths.
Increasing the number of interaction stages therefore increases the parameter count, making model compactness an important concern despite the simplified tracking pipeline.
\vspace{-10pt}
\paragraph{Efficient Visual Tracking.}
Efficient tracking methods address this model overhead through lightweight architecture design and model compression.
LightTrack~\citep{yan2021lighttrack} searches for compact tracking networks, while HiT~\citep{kang2023exploring} adapts lightweight hierarchical Vision Transformers for tracking.
For existing Transformer trackers, LiteTrack~\citep{wei2024litetrack} combines layer pruning with asynchronous feature extraction to reduce model size and redundant computation.
Depth reduction, however, leaves fewer stages for feature refinement.
To improve the accuracy of compact models, MixFormerV2~\citep{cui2023mixformerv2} combines progressive depth pruning with multi-stage distillation.
CompressTracker~\citep{hong2025general} employs replacement training and stage-wise feature supervision, while UETrack~\citep{kang2026uetrack} introduces target-aware adaptive distillation.
These distillation-based approaches improve compact trackers but require an additional teacher model and corresponding supervision during training.
\vspace{-10pt}
\paragraph{Looped Vision Transformers.}
Looped vision Transformers reuse parameters across multiple computation steps, allowing iterative feature refinement without assigning independent parameters to every step.
Recursive and single-block recurrent architectures have been explored for image recognition~\citep{shen2022sret,byra2026bvit}.
Related work also distills pretrained vision encoders into a few recurrently applied blocks to approximate their original multi-layer computation~\citep{jacobs2026block}.
Beyond recognition, shared-block recurrence has been applied to multi-view 3D reconstruction for iterative feature refinement~\citep{burzio2026dejaview}.

\section{Adapting Transformers for Recurrent Tracking}
\label{sec:method}
\begin{figure}[!t]
    \centering
    \includegraphics[width=\linewidth]{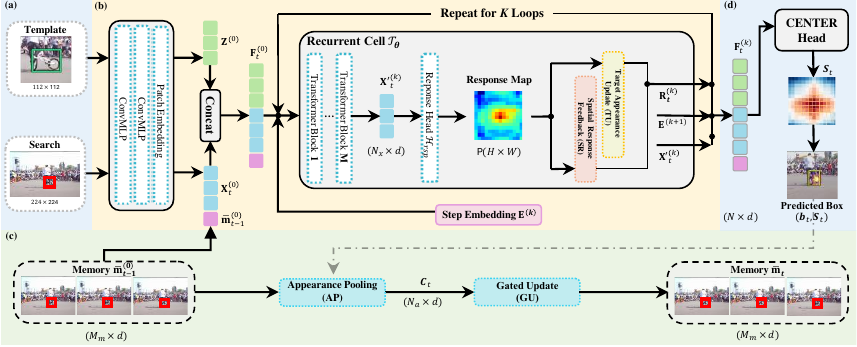}
    \caption{\textbf{Overview of LoopTrack.} (a) Input template and search region. (b) Target-Aware Looping (TAL) for recurrent feature refinement. (c) Gated Target Memory (GTM) for cross-frame target modeling. (d) Tracking head for target localization.}
    \label{fig:method}
    \vspace{-10pt}
\end{figure}

This section introduces our method. First, we introduce the baseline framework we adopt, namely the one-stream Transformer tracking framework. Subsequently, we propose LoopTrack, which is a parameter-efficient tracking framework proposed to address the large parameter overhead of existing Transformer trackers. On this basis, we further explore two lightweight designs for adapting to the recurrent structure, namely Target-Aware Looping (TAL) and Gated Target Memory (GTM). The overall framework of our method is shown in Fig.~\ref{fig:method}.

\subsection{Preliminaries}
\label{sec:preliminaries}

\paragraph{One-Stream Tracker.}
Our method builds upon a one-stream Transformer tracking
framework~\citep{ye2022joint,chen2025sutrack}, which jointly
performs feature extraction and template-search relation
modeling within a single backbone. Given an initial template image $z$ and a search region $x_t$ from frame $t$, a shared visual encoder produces template features $\mathbf Z_t$ and search features $\mathbf X_t$. The template and search features are concatenated along the token dimension and progressively refined through $L$ Transformer blocks:
\begin{equation}
\mathbf F_t = \mathcal T_{B_1}\cdots\mathcal T_{B_L}\!\left(\mathbf X_t,\mathbf Z_t\right).
\label{eq:one_stream_stack}
\end{equation}
where $\mathcal T_{B_\ell}$ denotes the $\ell$-th Transformer block with parameters $B_\ell$, for $\ell=1,\ldots,L$. The blocks in Eq.~\ref{eq:one_stream_stack} are applied sequentially from left to right, from block $1$ to block $L$, with no parameter sharing across blocks. Each block updates the template and search features through multi-head self-attention and feed-forward transformations~\citep{VaswaniSPUJGKP17}. The final search features $\mathbf X_t^{(L)}$ are extracted from the joint output $\mathbf F_t$ and passed to the prediction head to predict the target bounding box, $\hat{\mathbf b}_t=\mathcal H(\mathbf X_t^{(L)})$, where $\mathcal H$ denotes the prediction head together with bounding-box decoding.


However, in this stacked architecture, successive rounds of template-search interaction rely on separate parameter sets $B_1,\ldots,B_L$. Increasing the interaction depth therefore introduces additional block parameters and increases the model size. This coupling between interaction depth and parameter count limits the parameter efficiency
of conventional one-stream Transformer trackers.

\subsection{LoopTrack }
\label{sec:model_design}
To reduce the parameter overhead of independently parameterized Transformer blocks in the one-stream tracking framework described above, LoopTrack repeatedly applies a small set of Transformer blocks with shared parameters. This design enables multiple rounds of interaction between the template and search features while preserving the simplicity of the one-stream framework.
Specifically, LoopTrack is built on Fast-iTPN-T~\citep{tian2024fast}. Its patch embedding and first two ConvMLP stages form a shared visual encoder $\mathcal E$. Given an initial template image $z$ and the current search region $x_t$, the encoder extracts their features separately. Adding positional and token-type embeddings yields the template features $\mathbf Z^{(0)}\in\mathbb R^{N_z\times d}$ and search features $\mathbf X_t^{(0)}\in\mathbb R^{N_x\times d}$. Where, $N_z$ and $N_x$ denote the respective token counts, and $d$ is the feature dimension. The visual encoder operates outside the loop and provides the initial features for subsequent interactions.

For recurrent interaction, we retain $M$ Transformer blocks from the main stage of the pretrained model, forming a recurrent cell $\mathcal T_\theta$ sharing parameters across loops. We construct variants with $M=1$, $2$, and $3$, denoted LoopTrack$_{\text{One}}$, LoopTrack$_{\text{Two}}$, and LoopTrack$_{\text{Three}}$, respectively (Tab.~\ref{tab:model_variants}). The initial joint features are formed by concatenating the search and template features as $\mathbf F_t^{(0)}=[\mathbf X_t^{(0)};\mathbf Z^{(0)}]$. These features are fed into the recurrent cell for interaction and refinement through multi-head self-attention and feed-forward transformations. The updated features are fed back into the same recurrent cell in the next loop, enabling iterative feature refinement over $K$ loops.

Formally, the first loop uses the initial features, with $\bar{\mathbf X}_t^{(0)}=\mathbf X_t^{(0)}$ and $\bar{\mathbf Z}_t^{(0)}=\mathbf Z^{(0)}$. At loop $k$, the input search and template features are concatenated and fed into the shared recurrent cell:
\begin{equation}
\mathbf F_t^{(k)} = \mathcal T_\theta\!\left([\bar{\mathbf X}_t^{(k-1)};\bar{\mathbf Z}_t^{(k-1)}]\right), \qquad k=1,\ldots,K,
\label{eq:til_recurrence}
\end{equation}
The joint output is split along the token dimension as $\mathbf F_t^{(k)}=[\bar{\mathbf X}_t^{(k)};\bar{\mathbf Z}_t^{(k)}]$, yielding the updated search and template features. For $k<K$, these features are passed directly to the next loop. After $K$ loops, the final search features $\bar{\mathbf X}_t^{(K)}$ are passed to the prediction head to estimate the target bounding box in the current frame.

\paragraph{Target-Aware Looping.}
To further exploit target cues to guide recurrent interaction, we design a lightweight response head in TAL. TAL uses intermediate target localization responses to guide template-search interaction across loops, enabling progressive feature refinement. At loop $k$, the shared recurrent cell takes $\bar{\bar{\mathbf X}}_t^{(k)}$ and $\bar{\mathbf Z}_t^{(k-1)}$ as the search and template inputs. The search input includes a learned step embedding $\mathbf E^{(k)}$ that encodes the loop index. This embedding is broadcast across search positions. The joint output $\mathbf F_t^{(k)}$ is split along the token dimension into search features $\bar{\mathbf X}_t^{(k)}$ and template features $\bar{\mathbf Z}_t^{(k)}$.

For $1\le k<K$, a lightweight response head predicts the intermediate target localization response as $P_t^{(k)}=\sigma\!\left(\mathcal H_{\mathrm{rsp}}\!\left(\operatorname{LN}(\bar{\mathbf X}_t^{(k)})\right)\right)$. $\operatorname{LN}$ and $\sigma$ denote layer normalization and the sigmoid function, respectively. The normalized search features are reshaped into a spatial grid before entering the response head. The head consists of a $3\times3$ depthwise convolution, GELU activation, and a $1\times1$ convolution. It produces a single-channel response map and shares parameters across loops.

Spatial response (SR) uses the intermediate localization response to provide spatial guidance for the next loop. To highlight relative response differences across positions, SR subtracts the spatial mean from the response. This removes the response offset shared by all positions. A fully connected layer then projects the resulting response to the search feature dimension at each position, allowing the current target estimate to guide subsequent feature interaction.

However, spatial guidance alone does not explicitly provide the target appearance at these positions. To incorporate this information, target appearance update (TU) uses the response to aggregate $\bar{\mathbf X}_t^{(k)}$ and extract target appearance information. The aggregated appearance updates an intra-frame appearance state, which is preserved unchanged between TU updates and reused at the next TU update. The updated state is fused with local search features through a gate to generate appearance feedback. Subsequent interactions can thus use both the location and appearance of the current target estimate. Since appearance cues from successive recurrent steps are often highly correlated, TU is applied only at loops $2$, $4$, and $6$ when $k<K$ to reduce redundant feature feedback and unnecessary computation. For $1\le k<K$, the search feedback is
\begin{equation}
\mathbf R_t^{(k)} =
\begin{cases}
\operatorname{proj}\!\left(P_t^{(k)}-\operatorname{mean}(P_t^{(k)})\right)+\operatorname{gate}_{\mathrm{app}}\!\left(\bar{\mathbf X}_t^{(k)},P_t^{(k)}\right), & k\in\{2,4,6\},\\[4pt]
\operatorname{proj}\!\left(P_t^{(k)}-\operatorname{mean}(P_t^{(k)})\right), & \text{otherwise}.
\end{cases}
\label{eq:tal_feedback}
\end{equation}
$\operatorname{proj}$ denotes the fully connected projection, and $\operatorname{mean}$ computes the spatial mean. $\operatorname{gate}_{\mathrm{app}}$ includes target appearance aggregation, gated fusion, and feedback generation in TU. At other intermediate loops, only SR provides search feedback. The feedback and the next step embedding are added to the search output to form the next search input:
\begin{equation}
\bar{\bar{\mathbf X}}_t^{(k+1)} = \bar{\mathbf X}_t^{(k)}+\mathbf R_t^{(k)}+\mathbf E^{(k+1)}, \qquad 1\le k<K.
\label{eq:tal_search_update}
\end{equation}

The template output $\bar{\mathbf Z}_t^{(k)}$ is passed directly to the next loop. At each loop, the search and template inputs are concatenated and fed into the shared recurrent cell:
\begin{equation}
\mathbf F_t^{(k)} = \mathcal T_\theta\!\left([\bar{\bar{\mathbf X}}_t^{(k)};\bar{\mathbf Z}_t^{(k-1)}]\right), \qquad k=1,\ldots,K,
\label{eq:tal_recurrence}
\end{equation}
where $\mathbf F_t^{(k)}=[\bar{\mathbf X}_t^{(k)};\bar{\mathbf Z}_t^{(k)}]$. For each frame, the first loop uses $\bar{\bar{\mathbf X}}_t^{(1)}=\mathbf X_t^{(0)}+\mathbf E^{(1)}$ and $\bar{\mathbf Z}_t^{(0)}=\mathbf Z^{(0)}$. Search feedback is generated for $1\le k<K$ and incorporated into the search input of the next loop. After $K$ loops, the final search features $\bar{\mathbf X}_t^{(K)}$ are passed to the prediction head to estimate the target bounding box.

\paragraph{Gated Target Memory.}
While TAL uses intermediate localization responses to guide interaction within the current frame, it does not explicitly retain target appearance from previous frames. As the initial template image remains fixed during tracking,
historical observations can provide additional appearance information
for subsequent localization. To exploit this information and mitigate tracking drift, we introduce Gated Target Memory (GTM). GTM maintains a target memory (Mem) for recurrent interaction, extracts current appearance through appearance pooling (AP), and controls memory updates through a gated update (GU).

We initialize $\bar{\mathbf m}_0$ by applying coverage-weighted average pooling to the pre-loop template features using the initial target bounding box, adding a learnable memory embedding to the pooled features, and applying layer normalization. Each pooling weight is the fraction of the corresponding token's image region covered by the initial target bounding box. At frame $t$, the retained state initializes the current memory features as $\mathbf m_t^{(0)}=\bar{\mathbf m}_{t-1}$. To incorporate historical appearance into feature interaction, we concatenate these features with the search and template inputs and propagate them through the shared recurrent cell. With memory included, the complete recurrence becomes
\begin{equation}
\mathbf F_t^{(k)} = \mathcal T_\theta\!\left([\bar{\bar{\mathbf X}}_t^{(k)};\bar{\mathbf Z}_t^{(k-1)};\mathbf m_t^{(k-1)}]\right), \qquad k=1,\ldots,K,
\label{eq:looptrack_recurrence}
\end{equation}
with $\mathbf F_t^{(k)}=[\bar{\mathbf X}_t^{(k)};\bar{\mathbf Z}_t^{(k)};\mathbf m_t^{(k)}]$.
Within each loop, $\bar{\bar{\mathbf X}}_t^{(k)}$ follows the TAL formulation, while the template features are passed from the preceding loop, with $\bar{\mathbf Z}_t^{(0)}=\mathbf Z^{(0)}$. Joint self-attention allows historical target information in Mem to complement the localization feedback from TAL.

Although recurrent interaction refines the memory features within each frame, directly propagating $\mathbf m_t^{(K)}$ across frames may accumulate interaction-induced errors and redundant information. We therefore separate within-frame memory interaction from cross-frame memory updates. After $K$ loops, the CENTER head predicts the final center response $\mathbf s_t$ from $\bar{\mathbf X}_t^{(K)}$. Guided by this response, AP aggregates the cached pre-loop search features into an appearance candidate:
\begin{equation}
\mathbf c_t = \operatorname{LN}\!\left(\operatorname{pool}\!\left(\mathbf X_t^{(0)},\mathbf s_t\right)\right).
\label{eq:gtm_candidate}
\end{equation}
Specifically, $\operatorname{pool}$ selects positions in descending order of response until their cumulative normalized response reaches a prescribed threshold, then applies response-weighted average pooling. Restricting aggregation to likely target positions limits background contributions to the candidate. Moreover, using the final response for position selection and pre-loop features for aggregation separates localization guidance from appearance extraction, avoiding direct reuse of features already mixed with template and memory information.

Despite this selection, the appearance candidate remains dependent on localization accuracy. Background or distractor features introduced by an inaccurate prediction may contaminate the memory and affect subsequent frames. To reduce the influence of unreliable observations, GU derives a writing coefficient from the final response and regulates the fusion of the candidate with the memory:
\begin{equation}
\bar{\mathbf m}_t = \operatorname{LN}\!\left((1-\rho_t)\bar{\mathbf m}_{t-1}+\rho_t\mathbf c_t\right), \qquad \rho_t=\operatorname{gate}_{\mathrm{mem}}(\mathbf s_t).
\label{eq:gtm_update}
\end{equation}
To estimate $\rho_t\in[0,1]$, $\operatorname{gate}_{\mathrm{mem}}$ uses the peak response and its spatial concentration as cues for prediction reliability. These quantities describe response strength and spatial distribution, respectively. A larger coefficient increases the contribution of current appearance, whereas a smaller coefficient preserves more historical information. By adapting the writing strength to localization reliability, GU balances appearance adaptation with memory retention. Following the final localization, the updated memory initializes the next frame as $\mathbf m_{t+1}^{(0)}=\bar{\mathbf m}_t$.

\subsection{Learning Objective}
\label{sec:training}

Following the CENTER formulation~\citep{ye2022joint}, we supervise the final response with focal loss and the final box with $\ell_1$ and GIoU losses. For TAL, intermediate responses at loops $4$ and $6$ receive focal supervision, denoted by $\mathcal L_{\mathrm{rsp}}$. Auxiliary boxes use the same weighted regression objective ($\mathcal L_{\mathrm{box}}$), while localization quality is supervised with binary cross-entropy against the detached box IoU ($\mathcal L_{\mathrm{qual}}$). The auxiliary predictions enhance SR feedback during training and are disabled at inference, when feedback follows Eq.~\ref{eq:tal_feedback}.

For GTM, we supervise the memory writing coefficient $\rho_t$ with SmoothL1 loss against the detached IoU of the final prediction, denoted by $\mathcal L_{\mathrm{conf}}$. The overall objective is
\begin{equation}
\begin{aligned}
\mathcal L ={}& \mathcal L_{\mathrm{focal}}+\lambda_1\mathcal L_1+\lambda_{\mathrm{GIoU}}\mathcal L_{\mathrm{GIoU}}
&+\lambda_{\mathrm{rsp}}\mathcal L_{\mathrm{rsp}}+\lambda_{\mathrm{box}}\mathcal L_{\mathrm{box}}+\lambda_{\mathrm{qual}}\mathcal L_{\mathrm{qual}}+\lambda_{\mathrm{conf}}\mathcal L_{\mathrm{conf}}.
\end{aligned}
\label{eq:training_objective}
\end{equation}
We set $\lambda_1=5$, $\lambda_{\mathrm{GIoU}}=2$, $\lambda_{\mathrm{rsp}}=0.15$, $\lambda_{\mathrm{box}}=0.10$, and $\lambda_{\mathrm{qual}}=\lambda_{\mathrm{conf}}=0.05$.

\section{Experiments}
\label{sec:experiments}
\begingroup
\newcommand{\LTroom}[1]{\par\ifdim\pagetotal<\pagegoal\ifdim\dimexpr\pagegoal-\pagetotal\relax<#1\relax\newpage\fi\fi}
\setlength{\intextsep}{4pt}
\setlength{\columnsep}{10pt}
\setlength{\textfloatsep}{8pt}
\subsection{Implementation Details}
\label{sec:experimental_setup}
\noindent\paragraph{Model.}
LoopTrack is built on Fast-iTPN-T~\citep{tian2024fast}, which contains 24M parameters. We construct three variants with $M=1$, $2$, and $3$ Transformer blocks per recurrent cell, denoted LoopTrack$_{\text{One}}$, LoopTrack$_{\text{Two}}$, and LoopTrack$_{\text{Three}}$, respectively. Each cell is recurrently applied for $K$ loops with shared parameters, with TAL and GTM enabled in all variants. Tab.~\ref{tab:model_variants} summarizes the retained blocks, model size, and efficiency. Parameters and MACs cover the full tracker, and FPS is measured on an RTX 4090. We use PyTorch~\citep{paszke2019pytorch}, Python 3.12.12, and CUDA 12.6.

\begin{wraptable}[]{r}{0.47\textwidth}
\vspace{-6pt}
\centering
\captionsetup{
    position=top,
    skip=1pt,
    justification=justified,
    singlelinecheck=false
}
\caption{Model size and efficiency at $K=8$.}
\label{tab:model_variants}

\setlength{\tabcolsep}{1pt}

\resizebox{\linewidth}{!}{%
\begin{tabular}{llcccc}
\toprule
\rowcolor{LoopTrackHeader}
\textbf{Model} & \textbf{Blocks} & $M$
& \textbf{\shortstack{Params (M)}}
& \textbf{\shortstack{MACs (G)}}
& \textbf{FPS} \\
\midrule
LoopTrack$_{\text{One}}$   & $[6]$     & 1 & 3.4 & 3.79 & 153 \\
LoopTrack$_{\text{Two}}$   & $[5,6]$   & 2 & 4.9 & 6.70 & 119 \\
\rowcolor{LoopTrackHighlight}
LoopTrack$_{\text{Three}}$ & $[4,5,6]$ & 3 & 6.4 & 9.62 & 98 \\
\bottomrule
\end{tabular}%
}
\vspace{-4pt}
\end{wraptable}

\noindent\textbf{Training.}
We train LoopTrack on the splits of LaSOT~\citep{fan2021lasot}, GOT-10k~\citep{huang2021got10k}, TrackingNet~\citep{muller2018trackingnet}, and VastTrack~\citep{peng2024vasttrack}, sampling equally from four datasets.
We crop the template and search regions with factors of 2 and 4, respectively, and resize them to $112\times112$ and $224\times224$.
Following SUTrack~\citep{chen2025sutrack}, we initialize the retained backbone modules with ImageNet-pretrained weights.
During training, we vary the number of loops by sampling $u\sim\operatorname{Beta}(1,2)$ and setting $K=6+\operatorname{round}(6u)$, giving $K\in\{6,\ldots,12\}$. Each model is trained in two stages of 400 and 280 epochs, with 100,000 samples per epoch.
We use AdamW~\citep{loshchilov2019adamw} with a weight decay of $10^{-4}$ and a gradient clipping threshold of 0.1.
At the start of each stage, the learning rates are set to $10^{-5}$ for the encoder and $10^{-4}$ for the remaining modules, and both are reduced by a factor of 10 after 240 epochs within that stage. The second stage continues from the weights learned in the first stage, with the optimizer and learning rate scheduler reset. All models are trained on four RTX 4090 GPUs with a batch size of 128.

\noindent\textbf{Inference.}
We use a fixed number of loops, $K=8$, for all main comparisons. The auxiliary box and localization-quality branches are used only during training.

\subsection{Comparison with State-of-the-Art Trackers}
\label{sec:sota}

We evaluate LoopTrack on LaSOT~\citep{fan2019lasot}, LaSOT$_{\mathrm{ext}}$~\citep{fan2021lasot}, TrackingNet~\citep{muller2018trackingnet}, and GOT-10k~\citep{huang2021got10k}. As shown in Tab.~\ref{tab:main_sot}, compared with compact-design trackers, LoopTrack$_{\text{Three}}$ matches or exceeds UETrack-B~\citep{kang2026uetrack} across all tracking metrics while using $50.8\%$ fewer parameters. Compared with pruning-based methods, it surpasses LiteTrack-B9~\citep{wei2024litetrack} by $2.3$ SUC points on LaSOT while using only $6.4$M parameters compared with $54.9$M. Against compression-based trackers, LoopTrack$_{\text{Three}}$ further outperforms FARTrack-Tiny~\citep{wang2026fartrack} by $6.1$ SUC points under a comparable parameter budget ($6.4$M vs.~$6.8$M).

Notably, the same advantage remains evident in more compact LoopTrack variants. LoopTrack$_{\text{One}}$ achieves $66.2$ SUC on LaSOT with only $3.4$M parameters and a single Transformer block in its recurrent cell. It outperforms UETrack-T by $2.8$ points while using $43.3\%$ fewer parameters, demonstrating that loop reuse remains effective even under a highly constrained parameter budget. In addition, LoopTrack$_{\text{Two}}$, whose recurrent cell contains two Transformer blocks, reaches $68.4$ SUC with $4.9$M parameters, exceeding UETrack-S by $1.5$ points. Together, these results show that recurrent parameter sharing supports repeated feature refinement without a proportional increase in model parameters, leading to a favorable accuracy-parameter trade-off across different model scales.

\begin{table}[!t]
\centering
\captionsetup{skip=7pt}
\caption{Comparison with representative trackers on four benchmarks. Gray rows mark the method groups. The best score in each tracking metric is shown in \textbf{bold}, and LoopTrack$_{\text{Three}}$ is highlighted in light cyan. All tracking scores are percentages. GPU FPS values for other trackers are taken from the original papers. ``--'' denotes unreported results.}
\label{tab:main_sot}

\begingroup
\footnotesize
\setlength{\tabcolsep}{2.2pt}
\renewcommand{\arraystretch}{1.15}
\setlength{\extrarowheight}{0pt}

\setlength{\aboverulesep}{0pt}
\setlength{\belowrulesep}{0pt}

\resizebox{\textwidth}{!}{%
\begin{tabular}{l*{14}{c}}
\toprule

\rowcolor{LoopTrackHeader}
\textbf{Method}
& \multicolumn{2}{c}{\textbf{Efficiency}}
& \multicolumn{3}{c}{\textbf{LaSOT}}
& \multicolumn{3}{c}{\textbf{LaSOT$_{\mathrm{ext}}$}}
& \multicolumn{3}{c}{\textbf{TrackingNet}}
& \multicolumn{3}{c}{\textbf{GOT-10k}} \\

\noalign{%
  \begingroup
  \color{LoopTrackHeader}%
  \hrule height\cmidrulewidth
  \vskip-\cmidrulewidth
  \endgroup
}

\cmidrule(lr){2-3}
\cmidrule(lr){4-6}
\cmidrule(lr){7-9}
\cmidrule(lr){10-12}
\cmidrule(lr){13-15}

\rowcolor{LoopTrackHeader}
& \textbf{\shortstack{Params(M)}}
& \textbf{FPS}
& \textbf{SUC}
& {\boldmath$P_{\mathrm N}$}
& {\boldmath$P$}
& \textbf{SUC}
& {\boldmath$P_{\mathrm N}$}
& {\boldmath$P$}
& \textbf{SUC}
& {\boldmath$P_{\mathrm N}$}
& {\boldmath$P$}
& \textbf{AO}
& {\boldmath$\mathrm{SR}_{0.5}$}
& {\boldmath$\mathrm{SR}_{0.75}$} \\

\midrule



\tablegroup{Compression}

CompressTracker-4 \mbox{\citep{hong2025general}}
& 35.4 & 228
& 66.1 & 75.2 & 70.6
& 45.7 & -- & 50.8
& 82.1 & 87.6 & 80.1
& -- & -- & -- \\

CompressTracker-6 \mbox{\citep{hong2025general}}
& 49.8 & 162
& 67.5 & 77.5 & 72.4
& 46.7 & -- & 52.5
& 82.9 & 87.8 & 81.5
& -- & -- & -- \\

CompressTracker-8 \mbox{\citep{hong2025general}}
& 63.7 & 127
& 68.4 & 78.0 & 73.1
& 47.2 & -- & 53.1
& \textbf{83.3} & \textbf{88.0} & \textbf{81.9}
& -- & -- & -- \\

FARTrack-Pico \mbox{\citep{wang2026fartrack}}
& 2.8 & 343
& 58.6 & 67.1 & 59.6
& 41.8 & -- & --
& 75.6 & 81.3 & 70.5
& 62.8 & 72.6 & 50.9 \\

FARTrack-Nano \mbox{\citep{wang2026fartrack}}
& 4.6 & 210
& 61.3 & 69.7 & 64.1
& 43.8 & -- & --
& 79.1 & 84.5 & 75.6
& 69.9 & 81.2 & 61.4 \\

FARTrack-Tiny \mbox{\citep{wang2026fartrack}}
& 6.8 & 135
& 63.2 & 71.6 & 66.7
& 45.0 & -- & --
& 80.7 & 85.6 & 77.5
& 70.6 & 81.0 & 63.8 \\

\tablegroup{Pruning}

MixFormerV2-S \mbox{\citep{cui2023mixformerv2}}
& 16.2 & 325
& 60.6 & 69.9 & 60.4
& 43.6 & -- & 46.2
& 75.8 & 81.1 & 70.4
& 61.9 & 71.7 & 51.3 \\

LiteTrack-B4 \mbox{\citep{wei2024litetrack}}
& 26.2 & 315
& 62.5 & 72.1 & 65.7
& -- & -- & --
& 79.9 & 84.9 & 76.6
& 65.2 & 74.7 & 57.7 \\

LiteTrack-B9 \mbox{\citep{wei2024litetrack}}
& 54.9 & 171
& 67.0 & 77.0 & 72.7
& -- & -- & --
& 82.4 & 87.3 & 80.4
& 72.2 & 82.3 & 69.3 \\

\tablegroup{Compact Design}

HCAT \mbox{\citep{chen2022efficient}}
& -- & 195
& 59.3 & 68.7 & 61.0
& 40.6 & -- & --
& 76.6 & 82.6 & 72.9
& 65.1 & 76.5 & 56.7 \\

HiT-Base \mbox{\citep{kang2023exploring}}
& 42.1 & 175
& 64.6 & 73.3 & 68.1
& 44.1 & -- & --
& 80.0 & 84.4 & 77.3
& 64.0 & 72.1 & 58.1 \\

SMAT \mbox{\citep{gopal2024separable}}
& 3.8 & 158
& 61.7 & 71.1 & 64.6
& -- & -- & --
& 78.6 & 84.2 & 75.6
& 64.5 & 74.7 & 57.8 \\

AsymTrack-B \mbox{\citep{zhu2025two}}
& 3.4 & 197
& 64.7 & 73.0 & 67.8
& 44.6 & -- & --
& 80.0 & 84.5 & 77.4
& 67.7 & 76.6 & 61.4 \\

UETrack-T \mbox{\citep{kang2026uetrack}}
& 6.0 & 221
& 63.4 & 72.5 & 65.1
& 42.2 & 51.5 & 46.1
& 78.9 & 83.8 & 74.8
& 65.3 & 75.1 & 58.4 \\

UETrack-S \mbox{\citep{kang2026uetrack}}
& 9.0 & 183
& 66.9 & 76.1 & 70.7
& 47.9 & 58.4 & 53.9
& 81.4 & 86.3 & 78.8
& 71.1 & 81.2 & 67.0 \\

UETrack-B \mbox{\citep{kang2026uetrack}}
& 13.0 & 163
& 69.2 & 78.4 & 73.8
& 48.4 & 59.0 & 54.5
& 82.7 & 87.4 & 80.7
& 72.6 & 82.5 & 69.8 \\

\tablegroup{Ours}

\textbf{LoopTrack$_{\text{One}}$}
& 3.4 & 153
& 66.2 & 76.4 & 69.7
& 45.9 & 55.7 & 50.7
& 79.9 & 85.1 & 77.0
& 68.2 & 78.1 & 61.8 \\

\textbf{LoopTrack$_{\text{Two}}$}
& 4.9 & 119
& 68.4 & 78.6 & 73.8
& 48.3 & 58.4 & 54.0
& 81.7 & 86.6 & 79.5
& 71.9 & 81.8 & 67.8 \\

\rowcolor{LoopTrackHighlight}
\textbf{LoopTrack$_{\text{Three}}$}
& 6.4 & 98
& 69.3 & 79.2 & 74.7
& 49.5 & 60.1 & 55.5
& 82.7 & 87.4 & 80.8
& \textbf{73.1} & \textbf{82.8} & 69.9 \\

\bottomrule
\end{tabular}%
}

\endgroup
\end{table}

\subsection{Ablation and Analysis}
\label{sec:ablation}

In this section, we conduct a series of ablation studies on LaSOT to investigate the factors contributing to the effectiveness of LoopTrack. We use LoopTrack$_{\text{Three}}$ with $M=3$ as the default setting.

\LTroom{6\baselineskip}
\begin{wraptable}[]{r}{0.46\textwidth}
\vspace{-4pt}
\centering
\captionsetup{
    position=top,
    skip=4pt,
    justification=raggedright,
    singlelinecheck=false
}
\caption{Ablation of LoopTrack components.}
\label{tab:overall_component}

\setlength{\tabcolsep}{10pt}
\resizebox{\linewidth}{!}{%
\begin{tabular}{cccccc}
\toprule
\rowcolor{LoopTrackHeader}
& \textbf{TAL}
& \textbf{GTM}
& \textbf{SUC}
& {\boldmath$P_{\mathrm N}$}
& {\boldmath$P$} \\
\midrule
\ding{182}
& -- & --
& 65.8 & 75.5 & 70.8 \\
\ding{183}
& \ding{51} & --
& 68.8 & 78.6 & 74.0 \\
\rowcolor{LoopTrackHighlight}
\ding{184}
& \ding{51} & \ding{51}
& \textbf{69.3}
& \textbf{79.2}
& \textbf{74.7} \\
\bottomrule
\end{tabular}%
}
\end{wraptable}

\noindent\textbf{Ablation of LoopTrack components.}
To evaluate the contributions of TAL and GTM, we progressively introduce the two lightweight designs. As shown in Tab.~\ref{tab:overall_component}, incorporating TAL improves SUC from $65.8\%$ to $68.8\%$, a gain of $3.0$ percentage points (\ding{182}~vs.~\ding{183}).
This improvement indicates that guiding subsequent interactions with the current target estimate helps the shared Transformer blocks refine template and search representations more effectively across loops. Building on TAL, GTM introduces historical target appearance into recurrent interactions, complementing within-frame localization feedback with temporal information. This further improves SUC to $69.3\%$ (\ding{183}~vs.~\ding{184}), demonstrating the complementary benefits of inter-loop target guidance and cross-frame target information. The complete model gains $3.5$ percentage points over the looped baseline (\ding{182}~vs.~\ding{184}), demonstrating the complementary benefits of TAL and GTM for recurrent feature refinement.

\LTroom{10\baselineskip}
\begin{wraptable}{r}{0.44\textwidth}
\vspace{-5pt}
\centering
\captionsetup{
    position=top,
    skip=4pt,
    justification=raggedright,
    singlelinecheck=false
}
\caption{Ablation of components in TAL.}
\label{tab:tal_components}

\setlength{\tabcolsep}{10pt}
\resizebox{\linewidth}{!}{%
\begin{tabular}{cccccc}
\toprule
\rowcolor{LoopTrackHeader}
& \textbf{SR}
& \textbf{TU}
& \textbf{SUC}
& {\boldmath$P_{\mathrm N}$}
& {\boldmath$P$} \\
\midrule
\ding{182}
& -- & --
& 68.5 & 78.3 & 73.7 \\
\ding{183}
& \ding{51} & --
& 68.8 & 78.5 & 74.1 \\
\rowcolor{LoopTrackHighlight}
\ding{184}
& \ding{51} & \ding{51}
& \textbf{69.3}
& \textbf{79.2}
& \textbf{74.7} \\
\bottomrule
\end{tabular}%
}
\end{wraptable}

\noindent\textbf{Effectiveness of TAL design components.}
Tab.~\ref{tab:tal_components} evaluates SR and TU with GTM enabled in all configurations, while keeping the recurrent configuration and intermediate supervision fixed. SR improves SUC from $68.5\%$ to $68.8\%$ (\ding{182}~vs.~\ding{183}), showing that intermediate localization results provide effective spatial guidance for feature interaction in the next loop. The complete configuration achieves $69.3\%$ SUC. TU uses target appearance to guide search-feature updates. This suggests that updating search features with target information complements spatial guidance and improves recurrent interaction.

\noindent\textbf{Effectiveness of GTM components.}
We examine the components of GTM with TAL enabled in all configurations (Tab.~\ref{tab:gtm_components}). A static memory constructed from the initial template yields only a $0.1$-point SUC gain (\ding{182}~vs.~\ding{183}). Adding AP raises SUC to $69.1\%$ (\ding{184}), showing that recent target appearance provides subsequent frames with useful information beyond the initial template. Further introducing GU improves SUC to $69.3\%$ (\ding{185}), demonstrating the benefit of selectively updating the memory based on localization reliability. Together, AP and GU enable GTM to incorporate recent appearance while suppressing unreliable updates.

\LTroom{8\baselineskip}
\begin{wraptable}[]{r}{0.44\textwidth}
\vspace{-4pt}
\centering
\captionsetup{position=top,skip=1pt,justification=raggedright,singlelinecheck=false}
\caption{Ablation of components in GTM.}
\label{tab:gtm_components}

\setlength{\tabcolsep}{10pt}
\resizebox{\linewidth}{!}{%
\begin{tabular}{ccccccc}
\toprule

\rowcolor{LoopTrackHeader}
& \textbf{Mem}
& \textbf{AP}
& \textbf{GU}
& \textbf{SUC}
& {\boldmath$P_{\mathrm N}$}
& {\boldmath$P$} \\

\midrule

\ding{182}
& -- & -- & --
& 68.8 & 78.6 & 74.1 \\

\ding{183}
& \ding{51} & -- & --
& 68.9 & 78.7 & 74.2 \\

\ding{184}
& \ding{51} & \ding{51} & --
& 69.1 & 79.0 & 74.4 \\

\rowcolor{LoopTrackHighlight}
\ding{185}
& \ding{51} & \ding{51} & \ding{51}
& \textbf{69.3}
& \textbf{79.2}
& \textbf{74.7} \\

\bottomrule
\end{tabular}%
}
\end{wraptable}
To reduce the influence of unreliable updates on subsequent localization, GU controls memory updates according to prediction reliability, further improving SUC to $69.3\%$ (\ding{185}). These results show that, despite their simplicity, AP and GU effectively exploit cross-frame target information through target appearance selection and gated memory updates, further improving looped tracking performance.

\LTroom{7\baselineskip}
\begin{wraptable}[]{r}{0.44\textwidth}
\vspace{-2pt}
\centering
\setlength{\aboverulesep}{1pt}
\setlength{\belowrulesep}{1pt}
\captionsetup{position=top,skip=2pt,justification=raggedright,singlelinecheck=false}
\caption{Ablation of supervision in TAL.}
\label{tab:intermediate_prediction}

\setlength{\tabcolsep}{10pt}
\resizebox{\linewidth}{!}{%
\begin{tabular}{cccccc}
\toprule

\rowcolor{LoopTrackHeader}
& {\boldmath$\mathcal L_{\mathrm{box}}$}
& {\boldmath$\mathcal L_{\mathrm{qual}}$}
& \textbf{SUC}
& {\boldmath$P_{\mathrm N}$}
& {\boldmath$P$} \\

\midrule

\ding{182}
& -- & --
& 68.8 & 78.7 & 74.1 \\

\ding{183}
& \ding{51} & --
& 69.1 & 78.9 & 74.5 \\

\ding{184}
& -- & \ding{51}
& 69.0 & 79.0 & 74.3 \\

\rowcolor{LoopTrackHighlight}
\ding{185}
& \ding{51} & \ding{51}
& \textbf{69.3}
& \textbf{79.2}
& \textbf{74.7} \\

\bottomrule
\end{tabular}%
}
\end{wraptable}

\noindent\textbf{Effectiveness of auxiliary supervision in TAL.}
With GTM enabled in all configurations, we keep the TAL prediction branches and their feedback fixed during training and vary only the use of $\mathcal L_{\mathrm{box}}$ and $\mathcal L_{\mathrm{qual}}$ to examine the contribution of auxiliary supervision (Tab.~\ref{tab:intermediate_prediction}). Either loss improves tracking, and their combination raises SUC from $68.8\%$ to $69.3\%$ (\ding{182}~vs.~\ding{185}). This shows that supervising intermediate localization and its quality helps the model use target estimates to guide subsequent feature interaction. Auxiliary branches are disabled at inference, so these gains add no extra computation.

\noindent\textbf{Generality across Transformer trackers.}
To examine whether LoopTrack depends on the Fast-iTPN-T architecture used in our primary model, we further apply LoopTrack$_{\text{Three}}$ to UTPTrack-O-256~\cite{wu2026utptrack} and OSTrack-256~\citep{ye2022joint}. We use the same configuration with a three-block recurrent cell, $K=8$, TAL, and GTM, without tracker-specific architectural modification or hyperparameter tuning. Candidate elimination (CE) is disabled for OSTrack-256.

\begin{wraptable}[]{r}{0.48\textwidth}
\vspace{-4pt}
\centering
\captionsetup{
    position=top,
    skip=2pt,
    justification=justified,
    singlelinecheck=false
}
\caption{Generality of LoopTrack across different Transformer trackers on LaSOT.}
\label{tab:generality}

\setlength{\tabcolsep}{5pt}
\resizebox{\linewidth}{!}{%
\begin{tabular}{lcccc}
\toprule

\rowcolor{LoopTrackHeader}
\textbf{Variant}
& \textbf{\shortstack{Params (M)}}
& \textbf{SUC}
& {\boldmath$P_{\mathrm N}$}
& {\boldmath$P$} \\

\midrule

UTPTrack-O-256
& 92.5 & 67.40 & 76.30 & 72.50 \\

\rowcolor{LoopTrackHighlight}
w/ LoopTrack$_{\text{Three}}$
& 24.0 & 67.04 & 76.36 & 72.21 \\

\midrule

OSTrack-256 (w/o CE)
& 92.52 & 68.70 & 78.10 & 74.60 \\

\rowcolor{LoopTrackHighlight}
w/ LoopTrack$_{\text{Three}}$
& 24.1 & 68.26 & 77.82 & 74.21 \\

\bottomrule
\end{tabular}%
}
\end{wraptable}

As shown in Tab.~\ref{tab:generality}, LoopTrack reduces the parameter count of UTPTrack-O-256 and OSTrack-256 by $74.1\%$ and $74.0\%$, while the corresponding LaSOT SUC decreases are only $0.36$ and $0.44$ points. The precision metrics are also largely preserved. The consistent behavior across different Transformer trackers, obtained without tracker-specific tuning, supports the transferability of recurrent parameter sharing beyond the architecture used in our primary model. This suggests that LoopTrack can serve as a general parameter-efficient design for Transformer tracking rather than being tied to a specific backbone architecture. The remaining difference in absolute parameter count reflects the architecture-specific components retained by each host tracker.

\LTroom{10\baselineskip}
\begin{wrapfigure}[10]{r}{0.36\textwidth}
    \vspace{-\intextsep}
    \centering
    \captionsetup{
        width=\linewidth,
        skip=0pt,
        justification=centering,
        singlelinecheck=true
    }
    \includegraphics[
        width=\linewidth,
        trim=0bp 3bp 10bp 4bp,
        clip
    ]{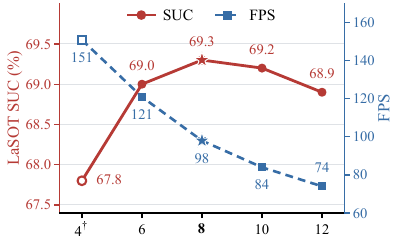}
    \caption{Effect of $K$.}
    \label{fig:depth_looptrack}
\end{wrapfigure}

\noindent\textbf{Flexible inference depth in LoopTrack.}
Fig.~\ref{fig:depth_looptrack} evaluates the same model at different loop counts $K$. Performance peaks at $K=8$ with $69.3$ SUC, while deeper inference provides no further gains. Despite lying outside the training range $[6,12]$, $K=4$ still achieves $67.8$ SUC at $151$ FPS without retraining, demonstrating generalization to unseen inference depths. These results show that recurrent parameter sharing enables a single LoopTrack model to support different accuracy-speed trade-offs. The optimum at $K=8$ is also consistent with the TU schedule ${2,4,6}$, as deeper inference adds recurrent refinement without further appearance updates.

\begin{wraptable}[]{r}{0.42\textwidth}
\vspace{-4pt}
\centering
\setlength{\aboverulesep}{1pt}
\setlength{\belowrulesep}{1pt}
\captionsetup{
    position=top,
    skip=2pt,
    justification=justified,
    singlelinecheck=false
}
\caption{GPU memory efficiency of LoopTrack, where \textbf{BS} denotes batch size.}
\label{tab:gpu_memory}

\setlength{\tabcolsep}{12pt}
\resizebox{\linewidth}{!}{%
\begin{tabular}{lcc}
\toprule

\rowcolor{LoopTrackHeader}
\textbf{Model}
& \textbf{BS = 1}
& \textbf{BS = 32} \\

\midrule

LoopTrack$_{\text{One}}$
& 46.09 & 368.28 \\

LoopTrack$_{\text{Two}}$
& 54.00 & 376.74 \\

\rowcolor{LoopTrackHighlight}
LoopTrack$_{\text{Three}}$
& 61.38 & 383.36 \\

\bottomrule
\end{tabular}%
}
\end{wraptable}

\noindent\textbf{Efficiency Analysis.}
To evaluate inference memory efficiency, we measure peak GPU memory for all three variants at $K=8$ (Tab.~\ref{tab:gpu_memory}). Memory usage ranges from 46.09--61.38 MiB at batch size 1 and remains below 384 MiB at batch size 32. These results demonstrate that LoopTrack supports recurrent feature interaction with low memory overhead, facilitating deployment on memory-constrained devices.

\section{Conclusion}
\label{sec:conclusion}

We propose LoopTrack, a parameter-efficient Transformer tracking framework that repeatedly applies a small set of shared Transformer blocks, enabling iterative template-search interaction with substantially fewer parameters. Target-aware looping (TAL) uses intermediate localization responses to guide progressive feature refinement, while gated target memory (GTM) incorporates compact cross-frame target cues to mitigate tracking drift. Experiments on four benchmarks demonstrate a favorable accuracy-parameter trade-off without additional teacher-based distillation, establishing LoopTrack as a simple yet strong baseline for parameter-efficient Transformer tracking.


\newpage
\subsection*{AI use statement}

Generative AI tools were used only for language polishing and table formatting. We did not use these tools to generate data or experimental results. We did not use them to make mathematical claims, design experiments, or implement methods. We also did not use them to analyze results or interpret research findings. The authors carefully reviewed, revised, and verified all AI-assisted text. The authors take full responsibility for the final content of this work.





\bibliography{reference.bib}
\bibliographystyle{iclr2027_conference}

\clearpage
\appendix
\section*{SUPPLEMENTAL MATERIAL}
For better understanding of this work, we offer additional details, analysis, and results as follow:
\begin{itemize}[leftmargin=*, labelsep=0.5em]
    \item \textbf{\ref{sec:supp_implementation} \;Implementation Details}
    
    Further details are provided for target appearance update in TAL, appearance pooling and gated memory update in GTM, as well as the auxiliary prediction branches and their supervision.
    
    \item \textbf{\ref{sec:supp_experiments} \;More Ablation Studies}
    
    Additional ablations investigate loop-count sampling during training, memory confidence supervision, appearance update schedules, and memory candidate construction, followed by analyses of model complexity and inference efficiency.
    
    \item \textbf{\ref{sec:supp_visualization} \;Qualitative Analysis}
    
    Qualitative comparisons across recurrent depths and design configurations illustrate how recurrent refinement, inter-loop target guidance, and cross-frame memory affect localization.
    
    \item \textbf{\ref{sec:limitation} \;Limitation}
    
    Finally, we discuss the limitation of our LoopTrack.
\end{itemize}

\section{Implementation Details}
\label{sec:supp_implementation}
This section elaborates on target appearance update (TU) in TAL, appearance pooling (AP) and gated update (GU) in GTM, and auxiliary supervision during training.

\subsection{Target-Aware Looping}
\label{sec:supp_tal_details}
For target appearance update (TU), $\operatorname{gate}_{\mathrm{app}}$ in Eq.~\ref{eq:tal_feedback} integrates response-weighted appearance aggregation with gated feature fusion. Omitting frame and loop indices for brevity, let $\mathbf h$ denote the internal appearance state, initialized to zero at the beginning of each frame and retained between TU updates. The response $P$ weights the normalized search features $\bar{\mathbf X}$ to update this state as
\begin{equation}
\mathbf h^{+}=(1-g)\mathbf h+g\frac{\sum_i P_i\operatorname{LN}(\bar{\mathbf X}_i)}{\sum_i P_i}.
\label{eq:supp_tu_update}
\end{equation}
The index $i$ denotes a search position, $\operatorname{LN}$ denotes layer normalization, and $\odot$ below denotes element-wise multiplication. Confidence gate $g\in[0,1]$ is computed from the difference between the response peak and spatial mean through a learned affine transformation followed by sigmoid. It controls the relative contributions of the current appearance observation and the retained state. Broadcasting $\mathbf h^{+}$ across search positions and combining it with local features yields
\begin{equation}
\operatorname{gate}_{\mathrm{app}}(\bar{\mathbf X},P)_i=g\,\mathbf w_i\odot\left(\mathbf h^{+}+\alpha\operatorname{LN}(\bar{\mathbf X}_i)\right).
\label{eq:supp_tu_feedback}
\end{equation}
In Eq.~\ref{eq:supp_tu_feedback}, $\alpha\in[0,1]$ is a learnable coefficient for local features, and $\mathbf w_i$ combines channel-wise gating, response-based spatial weighting, and overall scaling. Spatial weights are obtained by dividing each response by the spatial mean and clipping the ratio to $[0.25,4]$. Gate $g$ additionally modulates the feedback magnitude. The resulting output supplies the appearance term of $\mathbf R_t^{(k)}$ in Eq.~\ref{eq:tal_feedback} and thereby contributes to the next search input in Eq.~\ref{eq:tal_search_update}. The updated state is retained for subsequent TU updates within the frame.

\subsection{Gated Target Memory}
\label{sec:supp_gtm_details}
Appearance pooling (AP) specifies the pooling operation in Eq.~\ref{eq:gtm_candidate}. Given the final CENTER response $\mathbf s_t$, AP normalizes the scores over all search positions and selects them in descending response order. Selection stops when the cumulative normalized response reaches the prescribed threshold, set to $\mu=0.75$. This threshold determines the retained fraction of total response mass, rather than a fixed proportion of search positions. For the selected set $\Omega_t$, the appearance candidate is obtained by response-weighted pooling of the cached pre-loop search features:
\begin{equation}
\mathbf c_t=\operatorname{LN}\!\left(\frac{\sum_{i\in\Omega_t}s_{t,i}\mathbf X_{t,i}^{(0)}}{\sum_{i\in\Omega_t}s_{t,i}}\right).
\label{eq:supp_ap_pool}
\end{equation}
Normalization over $\Omega_t$ ensures that the selected response weights sum to one. Position selection uses the final localization response, while appearance aggregation uses $\mathbf X_t^{(0)}$, preserving the distinction between localization guidance and appearance extraction established in the main paper. The number of selected positions varies with the response distribution, whereas their aggregation always produces one appearance candidate.

Gated update (GU) incorporates $\mathbf c_t$ into the persistent memory through Eq.~\ref{eq:gtm_update}. The writing coefficient $\rho_t=\operatorname{gate}_{\mathrm{mem}}(\mathbf s_t)$ is determined from the response peak and spatial concentration, as specified in the main paper. It weights the current candidate by $\rho_t$ and the previous memory $\bar{\mathbf m}_{t-1}$ by $1-\rho_t$. Layer normalization of their weighted sum yields $\bar{\mathbf m}_t$, which initializes $\mathbf m_{t+1}^{(0)}$ in the next frame. Thus, the candidate produced by AP is used directly in GU, and the updated cross-frame memory remains a single token.

During inference, AP and GU use the final CENTER response before windowing for position selection and reliability estimation. A Hann window is applied separately when selecting the location for box decoding. This preserves the original response distribution used to compute both the appearance aggregation weights and the memory writing coefficient.

\subsection{Auxiliary Supervision}
\label{sec:supp_auxiliary_details}

\paragraph{Auxiliary prediction and feedback.}
As described in the main paper, TAL uses intermediate target estimates to improve SR during training. At the supervised loops, auxiliary box and localization-quality branches are constructed from the intermediate response features. Foreground and background template features are first aggregated according to target-box coverage, and their relative similarity to the search features is used to adjust the response logits. A spatial softmax over the adjusted response then produces pooling weights for obtaining a target representation, from which the auxiliary box and localization quality are predicted.

The predicted box is converted into a Gaussian guidance map according to its center and extent. The map is scaled by the predicted localization quality and a learned coefficient before being added to the response used by SR. TU and response supervision still use the original response. At inference, the auxiliary box and quality branches are removed, and SR follows Eq.~\ref{eq:tal_feedback}. Tabs.~\ref{tab:tal_components} and~\ref{tab:intermediate_prediction} separately evaluate target feedback under fixed intermediate supervision and auxiliary supervision under a fixed training feedback configuration; all results use this inference setting.

\paragraph{Auxiliary losses.}
Following the learning objective in the main paper, intermediate responses at loops $4$ and $6$ are supervised with focal loss, denoted by $\mathcal L_{\mathrm{rsp}}$. The corresponding auxiliary boxes use the same weighted regression objective as the final prediction, denoted by $\mathcal L_{\mathrm{box}}$, while localization quality is supervised with BCE against the detached IoU of the auxiliary box, denoted by $\mathcal L_{\mathrm{qual}}$.

For all three auxiliary objectives, the losses from loops $4$ and $6$ are combined with weights $0.25$ and $0.75$, respectively:
\begin{equation}
\mathcal L_{\ast}
=
0.25\,\mathcal L_{\ast}^{(4)}
+
0.75\,\mathcal L_{\ast}^{(6)},
\qquad
\ast\in\{\mathrm{rsp},\mathrm{box},\mathrm{qual}\}.
\label{eq:supp_auxiliary_weighting}
\end{equation}

These losses are then weighted by
$\lambda_{\mathrm{rsp}}$, $\lambda_{\mathrm{box}}$, and
$\lambda_{\mathrm{qual}}$ in Eq.~\ref{eq:training_objective}.
Since the training depth is sampled from $6$ to $12$, both supervised
loops are always reached. When $K=6$, the prediction at loop $6$
still receives supervision, but no feedback is applied afterward.
TU follows its scheduled updates at loops $2$, $4$, and $6$ whenever
a subsequent loop exists.

\paragraph{Memory confidence loss.}
For GTM, we follow the definition in the main paper and supervise the memory writing coefficient $\rho_t$ with SmoothL1 loss against the detached IoU of the final prediction, denoted by $\mathcal L_{\mathrm{conf}}$. This encourages a larger contribution from the current appearance when localization is reliable, while preserving more historical memory when the current prediction is uncertain.

\section{More Ablation Studies}
\label{sec:supp_experiments}

The following ablations examine how the design choices in LoopTrack support effective feature interaction under shared parameters. We study training with varying loop counts, the timing of within-frame appearance feedback, and the construction and reliability of cross-frame memory, followed by their parameter and inference costs. Unless otherwise specified, experiments use LoopTrack$_{\text{Three}}$ on LaSOT with $M=3$, $K=8$, TAL, and GTM. Light cyan indicates the default configuration, and bold denotes the highest tracking accuracy in each metric column, including ties. SUC, $P_{\mathrm N}$, and $P$ are reported as percentages; differences in these metrics are expressed in percentage points.

\subsection{Loop-count sampling}

Since LoopTrack reuses the same recurrent cell across loops, the loop count $K$ determines the amount of iterative feature refinement during training. We therefore examine whether training over varying loop counts benefits the shared recurrent cell compared with fixed-$K$ training, and further compare different sampling distributions for $K$.

\LTroom{8\baselineskip}
\begin{wraptable}[]{r}{0.48\textwidth}
\vspace{-4pt}
\centering
\captionsetup{
    position=top,
    skip=4pt,
    justification=justified,
    singlelinecheck=false
}
\renewcommand{\arraystretch}{1.10}

\caption{Ablation of loop-count sampling.}
\label{tab:supp_k_sampling}

\setlength{\tabcolsep}{3pt}
\resizebox{\linewidth}{!}{%
\begin{tabular}{lccc}
\toprule

\rowcolor{LoopTrackHeader}
\textbf{Training distribution}
& {\boldmath$K=6$}
& {\boldmath$K=8$}
& {\boldmath$K=12$} \\

\midrule

Fixed $K=8$
& 68.6 & 68.9 & 68.7 \\

Uniform $\{6,\ldots,12\}$
& \textbf{69.0} & 69.2 & 69.1 \\

$\operatorname{Beta}(2,1)$
& 68.8 & 69.1 & \textbf{69.2} \\

\rowcolor{LoopTrackHighlight}
$\operatorname{Beta}(1,2)$
& \textbf{69.0} & \textbf{69.3} & 68.9 \\

\bottomrule
\end{tabular}%
}
\end{wraptable}

As shown in Tab.~\ref{tab:supp_k_sampling}, training with varying loop counts consistently improves over fixed-$K$ training, supporting loop-count sampling for the shared recurrent cell. Beta$(2,1)$ favors larger loop counts and performs best at $K=12$, whereas Beta$(1,2)$ gives the best result at the default $K=8$ and ties the best result at $K=6$. We therefore adopt Beta$(1,2)$, which places more emphasis on moderate loop counts while retaining training coverage across $K\in\{6,\ldots,12\}$, matching LoopTrack's goal of flexible feature refinement under a compact parameter budget.

\LTroom{7\baselineskip}
\begin{wraptable}[6]{r}{0.48\textwidth}
\vspace{-4pt}
\centering
\captionsetup{
    position=top,
    skip=4pt,
    justification=justified,
    singlelinecheck=false
}
\renewcommand{\arraystretch}{1.10}
\setlength{\aboverulesep}{0pt}
\setlength{\belowrulesep}{0pt}

\caption{Effect of explicit memory confidence supervision in GTM.}
\label{tab:supp_memory_confidence}

\setlength{\tabcolsep}{3pt}
\resizebox{\linewidth}{!}{%
\begin{tabular}{lccc}
\toprule

\rowcolor{LoopTrackHeader}
\textbf{Memory confidence supervision}
& \textbf{SUC}
& {\boldmath$P_{\mathrm N}$}
& {\boldmath$P$} \\

\midrule

Without $\mathcal L_{\mathrm{conf}}$
& 69.0 & 77.8 & 73.7 \\

\rowcolor{LoopTrackHighlight}
With $\mathcal L_{\mathrm{conf}}$
& \textbf{69.3}
& \textbf{79.2}
& \textbf{74.7} \\

\bottomrule
\end{tabular}%
}
\end{wraptable}

\subsection{Memory confidence supervision}

GTM controls the contribution of the current appearance candidate to persistent memory according to localization reliability. To examine whether explicitly supervising this reliability is beneficial, we compare training with and without $\mathcal L_{\mathrm{conf}}$ while keeping all other settings unchanged. As described in the main paper, $\mathcal L_{\mathrm{conf}}$ supervises the memory writing coefficient $\rho_t$ using the IoU of the final prediction, encouraging reliable observations to contribute more to memory while preserving historical information when localization is uncertain. Adding $\mathcal L_{\mathrm{conf}}$ improves SUC from $69.0$ to $69.3$, $P_{\mathrm N}$ from $77.8$ to $79.2$, and $P$ from $73.7$ to $74.7$. These gains support the reliability-aware update in GTM, showing that explicitly supervising the writing coefficient helps regulate how much current appearance is incorporated into persistent memory. 

\LTroom{8\baselineskip}
\begin{wraptable}[]{r}{0.48\textwidth}
\vspace{-4pt}
\centering
\captionsetup{
    position=top,
    skip=4pt,
    justification=justified,
    singlelinecheck=false
}
\renewcommand{\arraystretch}{1.20}
\caption{Effect of TU update schedule.}
\label{tab:supp_target_schedule}

\setlength{\tabcolsep}{10pt}
\resizebox{\linewidth}{!}{%
\begin{tabular}{lccc}
\toprule

\rowcolor{LoopTrackHeader}
\textbf{TU update loops}
& \textbf{SUC}
& {\boldmath$P_{\mathrm N}$}
& {\boldmath$P$} \\

\midrule

$\{2\}$
& 68.9 & 78.8 & 74.2 \\

$\{2,4\}$
& 69.1 & 79.0 & 74.2 \\

$\{4,6\}$
& 69.0 & 78.9 & 74.3 \\

\rowcolor{LoopTrackHighlight}
$\{2,4,6\}$
& \textbf{69.3}
& \textbf{79.2}
& \textbf{74.7} \\

Every intermediate loop ($k<K$)
& 69.0 & 78.8 & 73.9 \\

\bottomrule
\end{tabular}%
}
\end{wraptable}

\subsection{Appearance update schedule}

TU is applied at selected loops to refresh target appearance during recurrent refinement. Since appearance cues from adjacent recurrent steps are often highly correlated, updating TU at every loop can introduce redundant feedback before the shared recurrent cell has sufficiently refined the representation. We therefore adopt a spaced update schedule in our default configuration and evaluate different choices while keeping SR and the inference depth fixed.

As shown in Tab.~\ref{tab:supp_target_schedule}, the spaced schedule $\{2,4,6\}$ achieves the best performance. Introducing TU early and refreshing it at later stages allows updated target appearance to guide progressive feature refinement, whereas delaying the updates to $\{4,6\}$ is less effective. Applying TU at every intermediate loop also reduces performance, supporting our motivation that adjacent appearance estimates contain redundant information. These results justify the use of $\{2,4,6\}$ as a sparse appearance-refresh schedule for TAL.

\subsection{Memory candidate construction}

GTM separates within-frame memory interaction from cross-frame memory updates. After the final localization, AP uses the response $\mathbf s_t$ to select likely target positions and aggregates their appearance into the candidate $\mathbf c_t$, which is subsequently written into the persistent memory through GU. We examine two design choices in this process: the feature representation used for appearance aggregation and the response-mass threshold used for position selection.

\paragraph{Feature source.}
To examine why AP uses the cached pre-loop search features in Eq.~\ref{eq:gtm_candidate}, we vary the feature source used to construct $\mathbf c_t$ while keeping the selected positions, response weights, candidate normalization, and GU unchanged.

\LTroom{8\baselineskip}
\begin{wraptable}[]{r}{0.48\textwidth}
\vspace{-4pt}
\centering
\captionsetup{
    position=top,
    skip=4pt,
    justification=justified,
    singlelinecheck=false
}
\renewcommand{\arraystretch}{1.10}

\caption{Ablation of candidate source.}
\label{tab:supp_memory_source}

\setlength{\tabcolsep}{8pt}
\resizebox{\linewidth}{!}{%
\begin{tabular}{lccc}
\toprule

\rowcolor{LoopTrackHeader}
\textbf{Candidate source}
& \textbf{SUC}
& {\boldmath$P_{\mathrm N}$}
& {\boldmath$P$} \\

\midrule

Final memory $\mathbf m_t^{(K)}$
& 68.9 & 78.7 & 74.2 \\

Selected final search $\bar{\mathbf X}_t^{(K)}$
& 69.0 & 78.8 & 74.6 \\

\rowcolor{LoopTrackHighlight}
Selected pre-loop search $\mathbf X_t^{(0)}$
& \textbf{69.3}
& \textbf{79.2}
& \textbf{74.7} \\

\bottomrule
\end{tabular}%
}
\end{wraptable}

As shown in Tab.~\ref{tab:supp_memory_source}, using the selected pre-loop search features achieves the best result, with $69.3$ SUC, compared with $69.0$ for the final search features and $68.9$ for the final memory token. The final search features have already interacted with the template and historical memory, while $\mathbf m_t^{(K)}$ has additionally participated in memory-conditioned recurrent interaction. Using either representation as new memory content therefore carries information already mixed during within-frame interaction into the cross-frame update. In contrast, Eq.~\ref{eq:gtm_candidate} uses the final response for refined localization guidance while extracting appearance from $\mathbf X_t^{(0)}$ before recurrent mixing. This separation provides current-frame appearance for memory update without directly reusing features already mixed with template and memory information.  

\paragraph{Response-mass threshold.}
AP selects positions in descending response order until their cumulative normalized response reaches $\mu$. To examine how much target evidence should be retained for appearance aggregation, we vary $\mu$ while keeping the feature source, GU, and single-token memory representation unchanged.

\LTroom{8\baselineskip}
\begin{wraptable}[]{r}{0.48\textwidth}
\vspace{-4pt}
\centering
\captionsetup{
    position=top,
    skip=4pt,
    justification=justified,
    singlelinecheck=false
}
\renewcommand{\arraystretch}{1.00}

\caption{Ablation of response-mass threshold.}
\label{tab:supp_mu_sensitivity}

\setlength{\tabcolsep}{10pt}
\resizebox{\linewidth}{!}{%
\begin{tabular}{lccc}
\toprule

\rowcolor{LoopTrackHeader}
\textbf{Response mass} {\boldmath$\mu$}
& \textbf{SUC}
& {\boldmath$P_{\mathrm N}$}
& {\boldmath$P$} \\

\midrule

0.50
& 68.8 & 78.6 & 74.0 \\

0.65
& 69.1 & 78.7 & 74.4 \\

\rowcolor{LoopTrackHighlight}
0.75
& \textbf{69.3}
& \textbf{79.2}
& \textbf{74.7} \\

0.85
& 68.7 & 76.7 & 73.9 \\

0.95
& 68.9 & 78.6 & 74.2 \\

\bottomrule
\end{tabular}%
}
\end{wraptable}

As shown in Tab.~\ref{tab:supp_mu_sensitivity}, $\mu=0.75$ achieves the best performance, reaching $69.3$ SUC. A smaller threshold retains only the strongest responses and can omit complementary target regions, whereas a larger threshold includes more low-response positions and increases the contribution of background or distractor features. The adopted threshold therefore balances target appearance coverage and background suppression. Since $\mu$ specifies cumulative response mass rather than a fixed number of positions, AP can adapt its spatial support to the current localization response while still producing a single compact appearance candidate.

\subsection{Model complexity and efficiency}

Since LoopTrack reduces model parameters through recurrent parameter sharing, the additional designs for target-aware refinement should preserve this parameter-efficient structure. We therefore measure the parameter and inference overhead introduced by TAL and GTM on LoopTrack$_{\text{Three}}$ with $M=3$ and $K=8$.

\LTroom{8\baselineskip}
\begin{wraptable}[]{r}{0.48\textwidth}
\vspace{-6pt}
\centering
\captionsetup{
    position=top,
    skip=4pt,
    justification=justified,
    singlelinecheck=false
}
\renewcommand{\arraystretch}{1.10}

\caption{Model complexity and inference efficiency of TAL and GTM.}
\label{tab:supp_module_overhead}

\setlength{\tabcolsep}{3pt}
\resizebox{\linewidth}{!}{%
\begin{tabular}{lccc}
\toprule

\rowcolor{LoopTrackHeader}
\textbf{Configuration}
& \textbf{\shortstack{Params (M)}}
& \textbf{\shortstack{MACs (G)}}
& \textbf{FPS} \\

\midrule

Plain shared loop
& 6.373 & 9.609 & 108 \\

+ TAL
& 6.377 & 9.618 & 101 \\

\rowcolor{LoopTrackHighlight}
+ GTM
& 6.379 & 9.620 & 98 \\

\bottomrule
\end{tabular}%
}
\end{wraptable}



As shown in Tab.~\ref{tab:supp_module_overhead}, adding TAL and GTM increases the parameter count from $6.373$M to only $6.379$M and MACs from $9.609$G to $9.620$G, corresponding to approximately $0.09\%$ and $0.11\%$ overhead, respectively. Meanwhile, SUC improves from $65.8$ to $69.3$ in Tab.~\ref{tab:overall_component}. The plain shared-loop baseline already performs repeated template--search interaction. This shows that the improvement is obtained without increasing the number or capacity of Transformer blocks. Instead, TAL and GTM provide intermediate localization cues and cross-frame target information to make the shared Transformer blocks perform more effective feature refinement, while preserving the compact parameterization of LoopTrack. The ablation in Tab.~\ref{tab:tal_components} further supports the benefit of target feedback under fixed intermediate supervision.

The additional target guidance reduces throughput from $108$ to $98$ FPS, while the model remains at $6.379$M parameters. Thus, TAL and GTM improve recurrent feature refinement with negligible additional model capacity, preserving the parameter-efficient nature of the shared-loop architecture. However, the small increases in parameters and MACs do not imply an equally small runtime cost.

\noindent\textbf{Flexible inference depth in LoopTrack.}
A direct benefit of recurrent parameter sharing is that LoopTrack can vary its inference depth by changing the number of loops without changing the model parameters. To examine this property, we evaluate the same trained LoopTrack$_{\text{Three}}$ checkpoint with different loop counts $K$, thereby varying the amount of recurrent feature refinement while keeping the stored model parameters fixed.

\LTroom{9\baselineskip}
\begin{wraptable}[]{r}{0.48\textwidth}
\vspace{-4pt}
\centering
\captionsetup{
    position=top,
    skip=3pt,
    justification=justified,
    singlelinecheck=false
}
\renewcommand{\arraystretch}{1.10}
\setlength{\aboverulesep}{0pt}
\setlength{\belowrulesep}{0pt}

\caption{Effect of inference depth.}
\label{tab:k_ablation}

\setlength{\tabcolsep}{10pt}
\resizebox{\linewidth}{!}{%
\begin{tabular}{ccccccc}
\toprule

\rowcolor{LoopTrackHeader}
& {\boldmath$K$}
& \textbf{Params (M)}
& \textbf{FPS}
& \textbf{SUC}
& {\boldmath$P_{\mathrm N}$}
& {\boldmath$P$} \\

\midrule

\ding{182}
& $4^{\dagger}$ & 6.4 & 151
& 67.8 & 78.0 & 72.5 \\

\ding{183}
& 6 & 6.4 & 121
& 69.0 & 78.9 & 74.3 \\

\rowcolor{LoopTrackHighlight}
\ding{184}
& 8 & 6.4 & 98
& \textbf{69.3}
& \textbf{79.2}
& \textbf{74.7} \\

\ding{185}
& 10 & 6.4 & 84
& 69.2 & 79.1 & 74.5 \\

\ding{186}
& 12 & 6.4 & 74
& 68.9 & 78.9 & 74.3 \\

\bottomrule
\end{tabular}%
}

\par\vspace{2pt}
\end{wraptable}

As shown in Tab.~\ref{tab:k_ablation}, the parameter count remains fixed at $6.4$M for all inference depths, while varying $K$ provides different accuracy-speed operating points with the same checkpoint. Increasing $K$ from $6$ to $8$ improves SUC from $69.0$ to $69.3$, with throughput decreasing from $121$ to $98$ FPS. Notably, $K=4^{\dagger}$ is outside the training-depth range $[6,12]$, yet the same model still achieves $67.8$ SUC at $151$ FPS without retraining. Compared with the default $K=8$, reducing the loop count to $K=4$ halves the number of recurrent cell executions while retaining strong tracking accuracy. This indicates that the shared recurrent cell can generalize beyond the refinement depths observed during training and remains effective at a shallower inference depth.

Further increasing $K$ to $10$ and $12$ introduces additional recurrent refinement and block executions, but does not further improve accuracy, yielding $69.2$ and $68.9$ SUC, respectively. Thus, tracking accuracy does not increase monotonically with recurrent computation alone. Under the current configuration, $K=8$ provides the highest accuracy, while smaller $K$ offers faster inference with exactly the same model parameters. These results demonstrate that recurrent parameter sharing decouples inference depth from the number of independently parameterized Transformer blocks, allowing a single trained LoopTrack model to operate at different accuracy-speed trade-offs.

\section{Qualitative Analysis}
\label{sec:supp_visualization}

We complement the quantitative ablations with qualitative examples of recurrent refinement, TAL, and GTM. Fig.~\ref{fig:supp_loop_evolution} evaluates different recurrent depths using the same trained checkpoint. Fig.~\ref{fig:supp_tal_variants} and~\ref{fig:supp_gtm_variants} compare separately trained models under otherwise identical settings. We focus on target localization, target extent, and failures caused by distractors or appearance changes.

\subsection{Iterative refinement across loops}

Fig.~\ref{fig:supp_loop_evolution} shows how predictions evolve with recurrent depth. In \textit{basketball-11}, early loops are attracted to a distractor region, while later loops recover the designated target. In \textit{yoyo-17} and \textit{drone-2}, the main change is in target extent: later loops suppress background regions or recover parts of the target missed at earlier depths. Similar corrections of target extent are observed in \textit{bottle-1} and \textit{pool-15}. These refinements are produced by repeatedly applying the same shared recurrent cell, rather than introducing additional independently parameterized Transformer blocks.

Increasing the recurrent depth does not always improve localization. Some predictions at loop~12 have similar or lower overlap than those at loop~8, despite the additional block executions. For example, IoU at frame~761 of \textit{bottle-1} decreases from $0.83$ at loop~8 to $0.73$ at loop~12. This agrees with the inference-depth results in Fig.~\ref{fig:depth_looptrack} and Tab.~\ref{tab:k_ablation}, where accuracy also stops improving beyond the default depth. Thus, the qualitative examples do not support a simple explanation in which more recurrent computation alone leads to better localization. Further interaction can revise an already accurate estimate without improving its alignment, suggesting that the usefulness of successive updates matters in addition to their number.

\subsection{Target guidance within the recurrent loop}

Fig.~\ref{fig:supp_tal_variants} compares separately trained models with and without TAL. Without TAL, recurrent interaction can drift toward target-irrelevant responses. In \textit{microphone-2}, the response is attracted to the person rather than the microphone. In \textit{zebra-10}, the tracker is confused by a similar instance. In \textit{airplane-15}, the main error is inaccurate target extent with substantial background included in the prediction. The additional \textit{gametarget-13} and \textit{helmet-13} examples similarly show better target localization and extent estimation with TAL. Despite template conditioning, visually prominent regions or similar objects can still provide competing evidence during feature interaction. Repeating this interaction does not by itself ensure that later loops focus on the designated instance or its correct spatial extent.

TAL uses the current target estimate to guide subsequent loops. SR provides spatial guidance from the intermediate localization response, while TU provides target appearance information. Together, these cues allow subsequent interaction to use an explicit intermediate estimate of where the target is and what it looks like, which can help resolve the competing evidence seen in these examples. This helps the shared recurrent cell maintain target-aware refinement across loops, consistent with the SR/TU ablation in Tab.~\ref{tab:tal_components}, while retaining the same recurrent-cell capacity.

\subsection{Target appearance across frames}

Fig.~\ref{fig:supp_gtm_variants} compares separately trained models with and without GTM. Because the initial template remains fixed during tracking, it cannot reflect all appearance changes observed in later frames. This is evident in \textit{hat-5} and \textit{pig-2}, where similar instances act as distractors, and in \textit{flag-2}, where the target undergoes substantial appearance and shape variation. In \textit{lion-12} and \textit{sheep-3}, predictions with GTM also remain better aligned with the target in the displayed later frames. Appearance changes can weaken correspondence with the fixed template, while similar instances make current-frame evidence ambiguous. Additional within-frame interaction alone does not supply the recent target observations needed to address this temporal mismatch.

GTM supplies recent target appearance through cross-frame target memory. AP selects current target evidence from the pre-loop search features using the final localization response, and GU regulates the memory update according to prediction reliability. Reliability-dependent writing is relevant because an incorrectly localized region may otherwise introduce distractor appearance into memory and affect subsequent frames. The resulting memory complements the fixed initial template and helps maintain target identity across frames. The qualitative behavior agrees with the GTM ablation in Tab.~\ref{tab:gtm_components} and the memory-candidate analysis in Tab.~\ref{tab:supp_memory_source}.

TAL and GTM therefore operate at different temporal scales: TAL guides recurrent refinement within a frame, whereas GTM carries target appearance across frames. Both provide target information to the same compact shared recurrent structure. This connects the qualitative results to the central design: parameter sharing preserves repeated feature interaction under a small parameter budget, while complementary target cues improve how that interaction is used for localization.

\section{Limitation}
\label{sec:limitation}

LoopTrack primarily focuses on parameter-efficient Transformer tracking through recurrent parameter sharing. Reusing a small set of shared Transformer blocks substantially reduces the number of stored model parameters, while recurrent feature refinement still requires repeated block execution. Therefore, the reduction in model parameters does not necessarily translate into a proportional reduction in computation. Nevertheless, the same trained model can operate at different inference depths by varying the number of loops without changing its parameters, providing multiple accuracy-speed operating points. Future work may further explore adaptive recurrent depth to allocate computation more selectively while retaining the compact shared-parameter design.

\begin{figure}[bp]
\centering
\includegraphics[width=\linewidth]{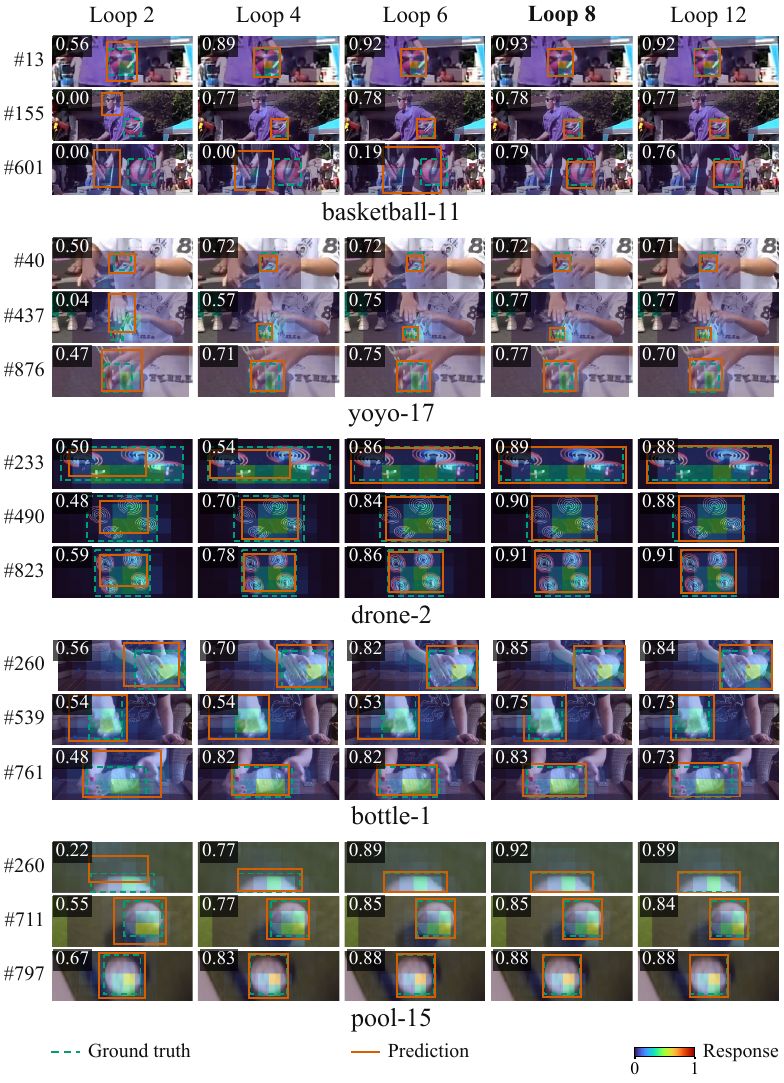}
\caption{Localization across recurrent depths using the same trained checkpoint. Columns correspond to loops $2$, $4$, $6$, $8$, and $12$. Green dashed boxes denote ground truth, orange boxes denote predictions, and heatmaps show spatial responses; values report IoU.}
\label{fig:supp_loop_evolution}
\end{figure}

\begin{figure}[!t]
\centering
\includegraphics[width=\linewidth]{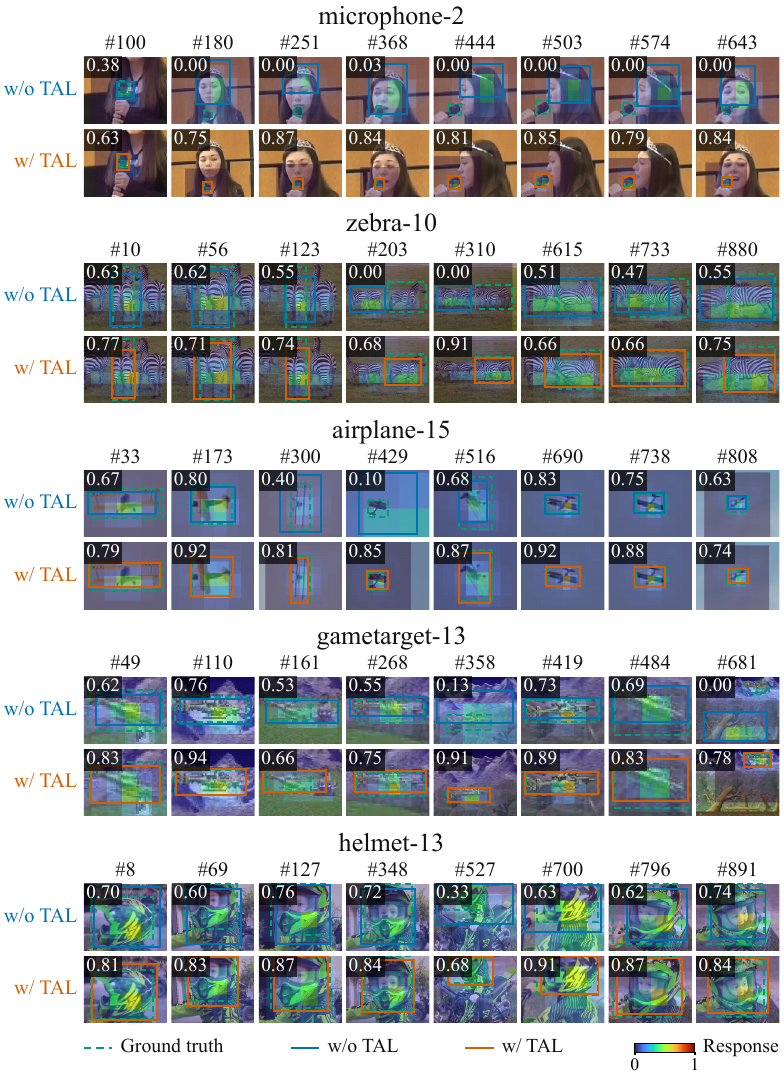}
\caption{Effect of TAL using separately trained models under identical settings. Upper and lower rows show results without and with TAL, respectively. Green dashed boxes denote ground truth, blue and orange boxes denote predictions, and heatmaps show spatial responses; values report IoU.}
\label{fig:supp_tal_variants}
\end{figure}

\begin{figure}[!t]
\centering
\includegraphics[width=\linewidth]{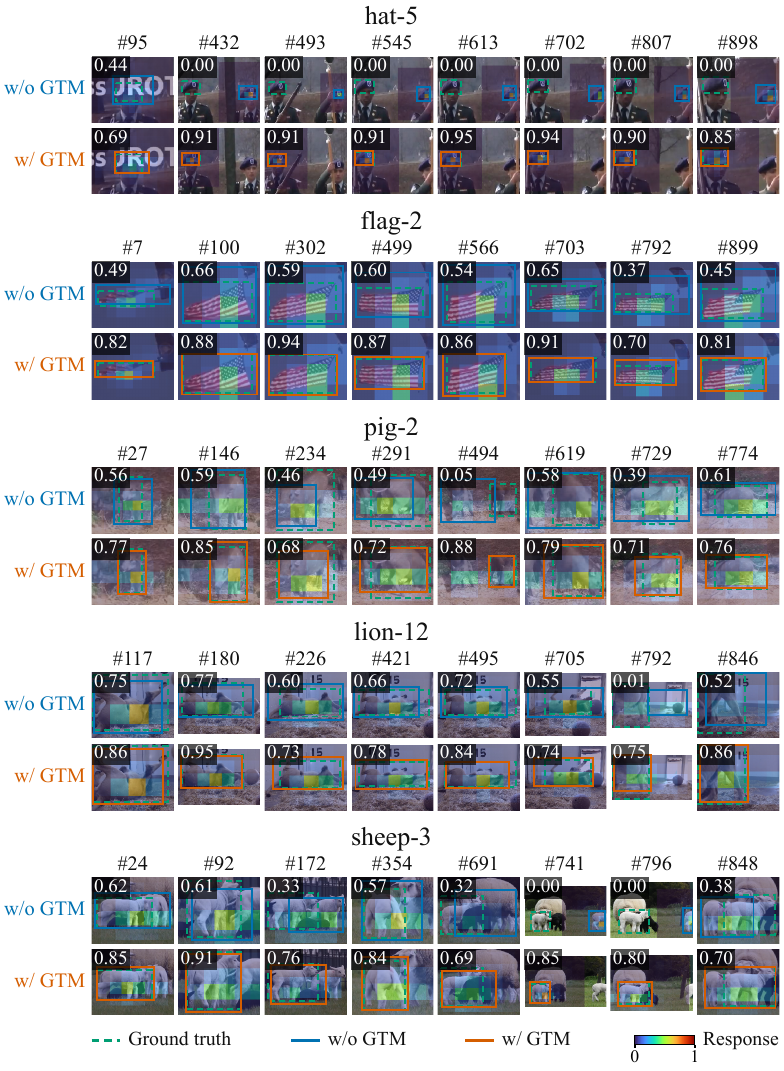}
\caption{Effect of GTM using separately trained models under identical settings. Upper and lower rows show results without and with GTM, respectively. Green dashed boxes denote ground truth, blue and orange boxes denote predictions, and heatmaps show spatial responses; values report IoU.}

\label{fig:supp_gtm_variants}
\end{figure}

\end{document}

%% file: math_commands.tex
\usepackage{amsmath,amsfonts,bm}

\def\eqref#1{equation~\ref{#1}}

\def\1{\bm{1}}

\DeclareMathAlphabet{\mathsfit}{\encodingdefault}{\sfdefault}{m}{sl}
\SetMathAlphabet{\mathsfit}{bold}{\encodingdefault}{\sfdefault}{bx}{n}

